\documentclass{article}
\usepackage{silence}
\PassOptionsToPackage{numbers}{natbib}

\usepackage[preprint]{neurips_2023}

\usepackage{longtable}
\usepackage[utf8]{inputenc} 
\usepackage[T1]{fontenc}   
\usepackage{hyperref}      
\usepackage{url}            
\usepackage{booktabs}       
\usepackage{amsfonts}       
\usepackage{CJKutf8}
\usepackage{nicefrac}       
\usepackage{microtype}      
\usepackage{xcolor}        
\usepackage{amsmath}
\usepackage{tcolorbox}
\usepackage{makecell} 
\usepackage{array}    
\usepackage{float}
\usepackage{graphicx}
\usepackage{ltablex}
\usepackage{lipsum}
\usepackage{graphicx}
\usepackage{wrapfig}
\usepackage[size=small]{caption}
\usepackage{mathtools}
\usepackage{tabularx}
\usepackage{amssymb}
\usepackage{amsthm}
\usepackage{soul}

\definecolor{textblue}{rgb}{.2,.2,.7}
\definecolor{textred}{rgb}{0.54,0,0}
\definecolor{textblack}{rgb}{0,0,0}
\definecolor{textgreen}{rgb}{0,0.53,0}
\usepackage{listings}
\lstdefinestyle{pythonstyle}{
    language=Python,
    morekeywords={None}, 
    breaklines=true,
}

\usepackage{array}
\newcolumntype{M}[1]{>{\centering\arraybackslash}m{#1}}

\title{AutoNeural: Co-Designing Vision–Language Models for NPU Inference}

\author{Wei Chen \\
Nexa AI\\
\texttt{alexchen@nexa.ai} \\
\And 
Liangmin Wu \\
Geely Auto\\
\texttt{Liangmin.Wu@geely.com} \\
\And
Yunhai Hu \\
Nexa AI\\
\texttt{yunhai@nexa.ai} \\
\And
Zhiyuan Li \\
Nexa AI\\
\texttt{zack@nexa.ai} \\
\And
Zhiyuan Cheng \\
Nexa AI\\
\texttt{perry@nexa.ai} \\
\And
Yicheng Qian \\
Nexa AI\\
\texttt{david@nexa.ai} \\
\And
Lingyue Zhu \\
Nexa AI\\
\texttt{alan@nexa.ai} \\
\And
Zhipeng Hu \\
Nexa AI\\
\texttt{victorhu@nexa.ai} \\
\And
Luoyi Liang \\
Geely Auto\\
\texttt{Luoyi.Liang@geely.com} \\
\And
Qiang Tang \\
Geely Auto\\
\texttt{Qiang.Tang2@geely.com} \\
\And
Zhen Liu \\
Geely Auto\\
\texttt{Zhen.Liu23@geely.com} \\
\And
Han Yang$^\dagger$ \\
Geely Auto\\
\texttt{Han.Yang6@geely.com} \\
}

\begin{document}
\raggedbottom
\begin{CJK*}{UTF8}{gbsn}
\renewcommand{\thefootnote}{\fnsymbol{footnote}}
\footnotetext[2]{Corresponding author}
\maketitle

\begin{abstract}
While Neural Processing Units (NPUs) offer high theoretical efficiency for edge AI, state-of-the-art Vision--Language Models (VLMs) tailored for GPUs often falter on these substrates. We attribute this hardware-model mismatch to two primary factors: the \textit{quantization brittleness} of Vision Transformers (ViTs) and the \textit{I/O-bound nature} of autoregressive attention mechanisms, which fail to utilize the high arithmetic throughput of NPUs. To bridge this gap, we propose \textbf{AutoNeural}, an NPU-native VLM architecture co-designed for integer-only inference. We replace the standard ViT encoder with a MobileNetV5-style backbone utilizing depthwise separable convolutions, which ensures bounded activation distributions for stable INT4/8/16 quantization. Complementing this, our language backbone integrates State-Space Model (SSM) principles with Transformer layers, employing efficient gated convolutions to achieve linear-time complexity. This hybrid design eliminates the heavy memory I/O overhead of Key-Value caching during generation. Our approach delivers substantial efficiency gains, reducing the quantization error of the vision encoder by up to 7$\times$ and end-to-end latency by 14$\times$ compared to conventional baselines. AutoNeural also delivers 3$\times$ faster decoding and a 4$\times$ longer context window than the baseline. We validate these improvements via a real-world automotive case study on the Qualcomm SA8295P SoC, demonstrating real-time performance for cockpit applications. Our results highlight that rethinking model topology specifically for NPU constraints is a prerequisite for robust multi-modal edge intelligence.
\end{abstract}

\section{Introduction}

Vision--language models (VLMs)~\cite{chen2024omnivlmtokencompressedsubbillionparametervisionlanguage, marafioti2025smolvlm, zhu2025internvl3, zhang2024vision,ghosh2024exploring} have fundamentally advanced the capability to jointly reason over images and text, achieving remarkable success in recognition, grounding, captioning, and visual instruction following. The dominant paradigm typically grafts a pretrained Vision Transformer (ViT) onto a decoder-only Large Language Model (LLM) via a projection layer, aligning the modalities through visual instruction tuning. While this GPU-centric design yields high accuracy at large resolutions, it introduces systemic inefficiencies when deployed on edge devices. Specifically, it suffers from: (i) high inference latency, dominated by the heavy computational cost of vision encoding and the autoregressive generation of the LLM (affecting both time-to-first-token and time-per-token); and (ii) inherent brittleness under low-precision execution, owing to the attention-heavy architectures in both vision and language components that are sensitive to quantization.

\paragraph{Why NPUs change the design space.}
Neural Processing Units (NPUs)~\cite{tan2021efficient,10472723,lee2021architecture} have emerged as the default compute substrate for mobile and edge intelligence, delivering high TOPS and energy efficiency through specialized integer operators and on-chip SRAM. However, state-of-the-art VLMs optimized for floating-point GPU execution often degrade significantly under NPU-friendly INT-$x$ quantization constraints. We identify two root causes for this hardware--software mismatch. First, Vision Transformers (ViTs)~\cite{khan2022transformers, chen2024comprehensive} exhibit \emph{quantization brittleness}---activations and weights in multi-head attention and RMSNorm~\cite{xu2019understanding} paths often possess outlier distributions that are sensitive to low-precision scaling, leading to sharp accuracy collapse at INT8/16. Second, Transformer-based language backbones suffer from \emph{memory I/O bottlenecks}. The autoregressive generation process necessitates repeated Key-Value (KV) cache access, creating memory-bound operations that saturate on-chip bandwidth. Consequently, the NPU's compute units stall despite high nominal TOPS, inflating latency and reducing end-to-end throughput.

\paragraph{Prefilling time dominates user-perceived latency.}
For interactive VLMs, system responsiveness is primarily governed by Time-To-First-Token (TTFT), which aggregates the latency of the vision encoder and the LLM prefilling phase across all input tokens. As input resolution or tiling strategies scale up to improve visual fidelity, the number of visual tokens increases quadratically, disproportionately elongating the prefilling phase. Consequently, the conventional strategy of scaling ViT width or depth yields diminishing returns on NPUs: marginal accuracy gains are negated by non-linear growth in TTFT and heightened sensitivity to quantization errors.

\paragraph{NPU-native co-design.}
We posit that achieving robust, low-latency multi-modal intelligence at the edge necessitates an \emph{NPU-native co-design} approach across two dimensions:
(1) \textbf{Topology}: Replacing global attention mechanisms with operator- and memory-efficient components that naturally stabilize INT-$x$ inference and minimize memory traffic;
(2) \textbf{Runtime}: Implementing scheduling strategies that respect NPU buffer hierarchies and bandwidth limits, constraining peak activation footprints and KV cache overhead to optimize both TTFT and generation speed.

\paragraph{Our approach.}
We introduce \textbf{AutoNeural}, an NPU-native VLM architecture that fundamentally rethinks vision encoding and language modeling for edge efficiency. For the vision encoder, we depart from standard ViTs in favor of a MobileNet-style architecture utilizing depthwise separable convolutions. This design eliminates the quadratic complexity and quantization brittleness of global attention, providing strong inductive biases for feature extraction while maintaining bounded activation distributions that are inherently stable for INT8/16 inference. For the language backbone, we adopt a hybrid Transformer-SSM architecture (inspired by Liquid AI principles), which interleaves Transformer layers with efficient gated convolutions based on structural state-space models. This topology offers linear-time complexity and compact state representations, obviating the need for explicit KV caching and reducing memory I/O by up to 60\% during generation. To our knowledge, this is the first framework to unify convolution-based visual stability with linear-complexity state-space modeling, effectively decoupling multi-modal reasoning capability from the heavy computational burden of global attention. We couple this architecture with an NPU-aware training recipe that integrates quantization-aware fine-tuning (QAT)~\cite{li2023loftq, bondarenko2024low, chen2025efficientqat}, mixed-precision constraints, and hardware-aligned calibration.

\paragraph{Key findings.}
Our architecture yields up to \(\text{7}\times\) lower quantization error and \(\text{14}\times\) lower end-to-end latency versus ViT-Transformer baselines under the same NPU precision constraints. The MobileNet encoder contributes to stable INT8/16 inference with minimal quantization degradation, while the Liquid AI backbone reduces memory I/O by up to 60\% during autoregressive generation. Hardware measurements on a commercial automotive SoC (Qualcomm SA8295P NPU) show real-time responsiveness in an in-car assistant scenario (Figure~\ref{fig:model_functions}), validating that both TTFT and time-per-token can be dramatically reduced without sacrificing accuracy. Ablations isolate the contribution of the MobileNet encoder, SSM integration, and token budgeting to the observed latency–accuracy Pareto improvements.

\paragraph{Contributions.}

In conclusion, this work makes three primary contributions to enable efficient vision–language modeling on NPUs:
\begin{itemize}
    \item We propose a novel \textbf{NPU-native architecture} combining a depthwise-convolutional vision encoder with a hybrid Transformer-SSM language backbone. This design solves the twin challenges of quantization brittleness and memory bottlenecks, outperforming traditional ViT-LLM pairs in INT8/16 robustness and efficiency.
    \item We introduce a proprietary \textbf{automotive-specific dataset} comprising 200k annotated samples. This dataset covers critical cockpit AI tasks---including driver monitoring, vehicle security, and parking localization---spanning diverse demographics and environmental conditions to benchmark domain-specific performance.
    \item We develop a comprehensive \textbf{NPU-aware training framework}, integrating QAT and calibration procedures specifically designed to minimize post-quantization drift and align vision--language representations for deployment on the Qualcomm SA8295P platform.
\end{itemize}

\begin{figure}[t]
\centering
\includegraphics[width=\textwidth]{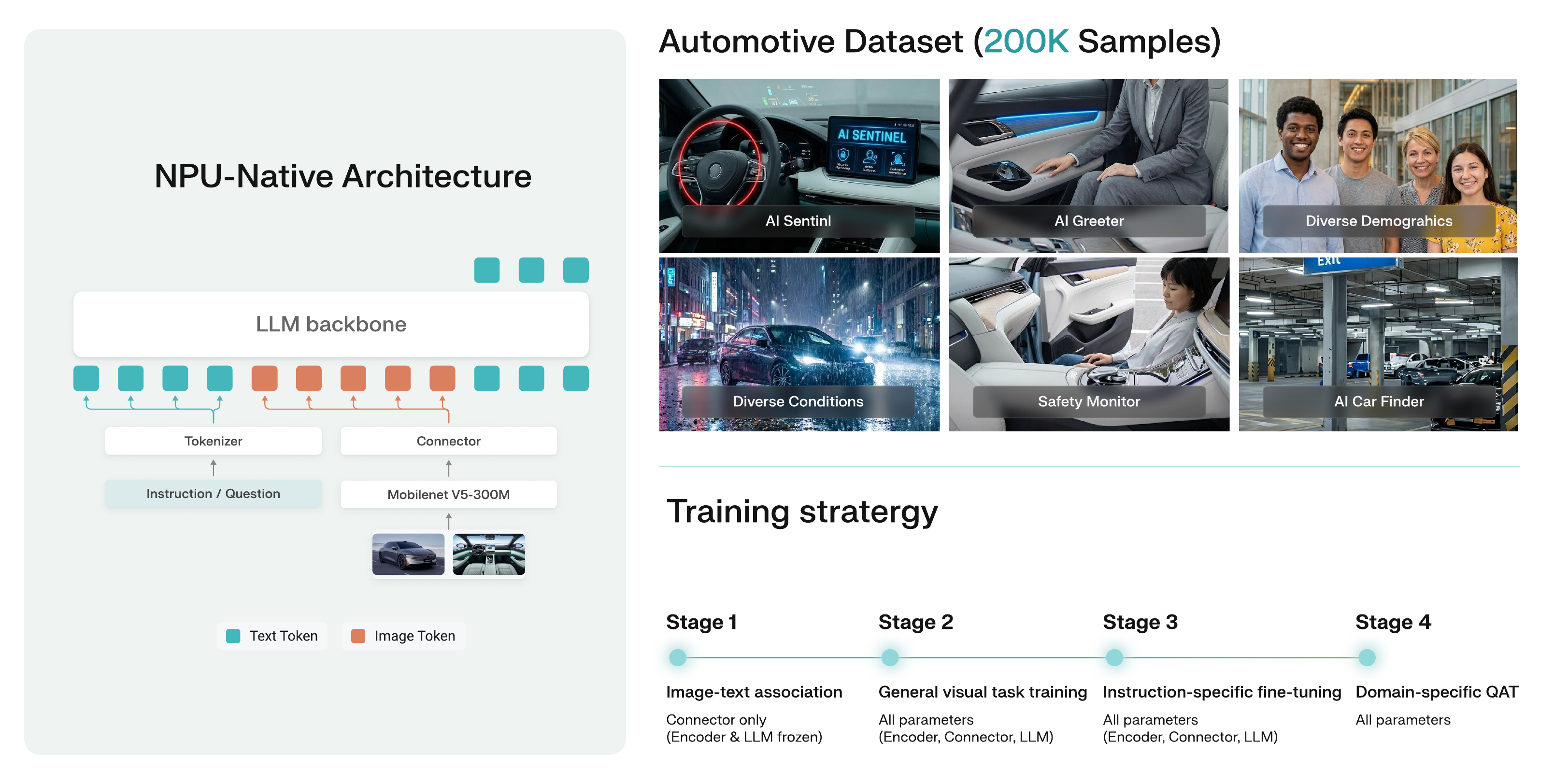}
\caption{Architecture overview of AutoNeural. The model comprises: (1) a MobileNet-based vision encoder with Multi-Scale Fusion Adapter (MSFA) that processes $768{\times}768$ images into 256 visual tokens, (2) a lightweight two-layer MLP connector without normalization for NPU quantization robustness, and (3) the Liquid AI 1.2B hybrid backbone with 16 layers that interleaves 10 gated-convolution layers with 6 Transformer attention layers to reduce memory I/O.}
\label{fig:model_arch}
\end{figure}

\section{Related Work}

Our work sits at the intersection of three research areas: multimodal models for automotive applications, NPU-optimized inference, and efficient vision encoders for on-device deployment. 

\paragraph{Multimodal models for automotive deployment.}
The automotive industry has increasingly adopted vision–language models to enable advanced driver assistance systems (ADAS), in-car assistants, and autonomous driving capabilities~\cite{guo2024vlm,huang2024vlm}. Recent work has explored the application of large multimodal models in automotive contexts, ranging from scenario generation to real-time perception. For instance, multimodal LLMs~\cite{chen2024octopusv3technicalreport, chen2024omnivlmtokencompressedsubbillionparametervisionlanguage} have been applied to autonomous driving tasks such as scenario generation and perception, demonstrating the potential of integrating visual and textual reasoning for improved situational awareness~\cite{chen-etal-2025-octopus,chen2024octopusv2ondevicelanguage,chen2024octopus,chen2024octoplannerondevicelanguagemodel,chen2024squidlongcontextnew}. However, most of these systems rely on GPU-based inference and have not been optimized for the power and latency constraints of in-vehicle NPUs. The deployment of such models in production vehicles requires careful co-design with edge hardware, a gap our work directly addresses. While existing automotive multimodal systems often compromise on model capacity or resolution to meet latency budgets, our MobileNet-Hybrid Transformer-SSM architecture demonstrates that hardware-aware design enables both high accuracy and low latency.

\paragraph{Models on NPU architectures.}
Neural Processing Units from vendors such as Qualcomm~\cite{zhu2025edge,xu2025fast}, MediaTek~\cite{park2023overviewing}, and Apple have become the primary compute substrate for on-device AI across mobile and automotive platforms. Recent efforts have focused on adapting large multimodal models for NPU deployment through quantization~\cite{wei2024advances, tmamna2024pruning}, pruning~\cite{dantas2024comprehensive}, and operator fusion~\cite{salmani2025llm}. For example, work on Qualcomm platforms~\cite{federici2024efficient, rajaee2025locallookaheadguidanceverifierintheloop, hegde2025distillingmultimodallargelanguage, farhadzadeh2025loraxbridgingfoundationmodels} has demonstrated that quantization-aware training and low-rank factorization can accelerate inference while maintaining acceptable accuracy. Similarly, frameworks such as MindVL~\cite{chen2025mindvl} have explored efficient training pipelines on Ascend NPUs, integrating distributed data loading and system-level scheduling. MiniCPM-V~\cite{yu2025minicpm,yao2024minicpm} represents another line of work targeting edge deployment through aggressive model compression, achieving strong performance on mobile devices. However, these approaches primarily focus on post-hoc optimization of GPU-first architectures rather than fundamental architectural redesign. In contrast, our work rethinks both the vision encoder and language backbone to match NPU operator sets and memory hierarchies, yielding more substantial latency and efficiency gains.

\paragraph{Efficient vision encoders for on-device deployment.}
The choice of vision encoder critically impacts both accuracy and deployment efficiency in multimodal systems. While Vision Transformers have dominated recent VLM architectures due to their strong performance, their global attention mechanisms and quantization brittleness pose significant challenges for on-device deployment. Recent work has explored alternative encoder designs optimized for edge scenarios. MobileNet architectures~\cite{chen2018enhanced, sinha2019thin, qin2018fd, khasoggi2019efficient}, originally designed for efficient image classification, employ depthwise separable convolutions that reduce parameter count and computation while maintaining strong inductive biases for visual features. Notably, PaliGemma~\cite{beyer2024paligemma, steiner2024paligemma}, a recent 3B-parameter VLM from Google, demonstrates that carefully designed smaller vision encoders can achieve competitive transfer performance. Beyond vision encoders, recent advances in sequence modeling have introduced State Space Models (SSMs)~\cite{gu2021efficiently, gu2024mamba, dao2024transformersssmsgeneralizedmodels, gu2021combining} as efficient alternatives to Transformers for language modeling. SSMs achieve linear-time complexity through selective state spaces and eliminate the need for explicit key-value caching, making them particularly attractive for memory-constrained NPU deployment. Building on these principles, Liquid Foundation Models~\cite{wu2024liquid, smekal2024towards, poli2403mechanistic} demonstrate efficient sequence modeling architectures that combine linear-time complexity with strong representational capacity through liquid structural state-space modeling. Our work adopts the Liquid AI architecture as our language backbone and systematically pairs it with a MobileNet-based encoder optimized for NPU-aware quantization. By replacing global attention in both the vision and language components, we address the dual bottlenecks of quantization brittleness and memory I/O that plague existing VLM architectures on NPUs.

\section{Methodology}

We present the architectural design and training protocol of our NPU-native vision–language model. Our approach consists of two key components: (1) a MobileNet-based vision encoder for efficient visual feature extraction; and (2) a hybrid Transformer-SSM language backbone that reduces memory I/O during autoregressive generation.

\subsection{Model Architecture}
\paragraph{Vision encoder}
Our vision encoder builds on MobileNetV5, exploiting depthwise separable convolutions for efficient feature extraction with bounded activations, and is initialized from the Gemma 3n-E4B vision checkpoint to align with the multimodal design of Gemma 3n~\cite{gemma_3n_2025,team2025gemma}. It consumes $768{\times}768$ inputs and produces a spatial feature map that is flattened into visual tokens. In contrast to ViTs with global self-attention, we employ a hierarchical stack of inverted residual (IR) blocks that naturally yields multi-scale features while preserving local receptive fields; this local-first design reduces quantization sensitivity, as convolution and normalization maintain more stable activation ranges than LayerNorm and multi-head attention, improving INT8/16 robustness. The network begins with a $3{\times}3$ stride-2 stem, followed by four stages in which stride-2 IR blocks downsample and stride-1 IR blocks refine. To inject long-range context at bounded cost, the late, low-resolution stages interleave sparse multi-query attention (MQA) bottlenecks. We aggregate information via a \emph{Multi-Scale Fusion Adapter} (MSFA): the last two stage outputs are upsampled to the finest tapped resolution, channel-concatenated, and processed by a universal inverted residual layer (pointwise$\rightarrow$depthwise$\rightarrow$pointwise) with \texttt{GELU} (tanh) activations and RMSNorm; a $3{\times}3$ stride-3 average pool yields a fused $16{\times}16{\times}2048$ tensor. This spatial feature map is then flattened into $256$ visual tokens, where each of the $16{\times}16$ spatial locations becomes a token with $2048$-dimensional features, balancing fidelity and LLM prefill cost while preserving accuracy and latency benefits.

\paragraph{Vision--language connector}
The vision--language connector serves as the alignment bridge between the visual encoder's output space and the language model's embedding space, transforming the visual token representations produced by the MobileNet encoder into the token embeddings expected by the language model backbone. 

Our connector architecture employs a lightweight two-layer multi-layer perceptron (MLP) with a GELU (Gaussian Error Linear Unit) activation between the two layers. This design deviates from the Gemma 3n-E4B projector, which incorporates two RMSNorm layers within the projection path. We deliberately avoid normalization layers in the connector to maintain NPU quantization robustness. Contemporary NPU quantization pipelines rely on static per-tensor quantization, where activation ranges must be determined at calibration time and fixed during deployment. RMSNorm introduces significant quantization challenges: its dynamic scaling operation—computing the root mean square over features at each position—produces activations with input-dependent distributions that are difficult to calibrate accurately with static ranges. 

\paragraph{Language model backbone}
For the language model, we adopt the Liquid AI 1.2B parameter architecture, a hybrid design that strategically interleaves Transformer self-attention layers with efficient sequence modeling layers based on liquid structural state-space principles. The sequence modeling layers implement gated convolutions with depthwise short-kernel operations, providing linear-time complexity during inference while maintaining compact state representations that avoid explicit key-value cache overhead. The Liquid AI architecture employs a 5:3 ratio with 16 total layers—10 layers use gated-convolution sequence modeling while 6 layers use Transformer self-attention. This design is particularly well-suited for NPU deployment for two reasons: (1) the convolution-based layers provide linear-time complexity during autoregressive generation, eliminating the quadratic scaling of attention and dramatically reducing memory footprint, and (2) the preserved Transformer layers maintain strong in-context learning and reasoning capabilities essential for vision–language tasks. The gated-convolution layers maintain a rolling state cache that operates with bounded memory during inference, requiring only a fixed-size context window rather than the full key-value history. During generation, this hybrid design reduces memory bandwidth pressure on the NPU by up to 60\% compared to a pure Transformer baseline, as the majority of layers avoid explicit KV cache reads/writes. Each sequence modeling layer is followed by a feed-forward network with SwiGLU activation, with RMSNorm applied before both types of layers to stabilize training and maintain activation distributions suitable for quantization. The complete architecture is illustrated in Figure~\ref{fig:model_arch}.

\subsection{Automotive dataset}

We collected and labeled 0.2M samples for intelligent cockpit tasks from scratch. Approximately 400 volunteers of different ages, genders, and skin tones participated in the data collection process to improve the generalization performance of the dataset. After data collection, we used a combination of automated and manual annotation to label the data accurately. Specifically, the dataset comprises the following task types: (1) AI Sentinel (56K samples): after the vehicle is turned off and locked, the camera monitors the vehicle's surroundings in real time and identifies destructive external behaviors, such as scratching, prying, or spray-painting the vehicle, providing 24-hour real-time protection; (2) AI Greeter (50K samples): when an acquaintance approaches the car, their identity is recognized and confirmed to assist them in unlocking the car or opening the trunk in advance, offering a personalized welcome experience; (3) AI Car Finder (44K samples): identifies salient cues related to the vehicle's location in a parking lot, such as the floor, zone, parking space, and surrounding vehicles, enabling the owner to find the vehicle quickly and accurately; (4) Safety when Exiting or Starting the Car (50K samples): when passengers exit or the car starts, the system identifies potential hazards around the vehicle, such as obstacles, pedestrians, or small animals, and provides safety reminders to the user.

\subsection{NPU deployment validation}

\begin{figure}[t]
\centering
\includegraphics[width=\textwidth]{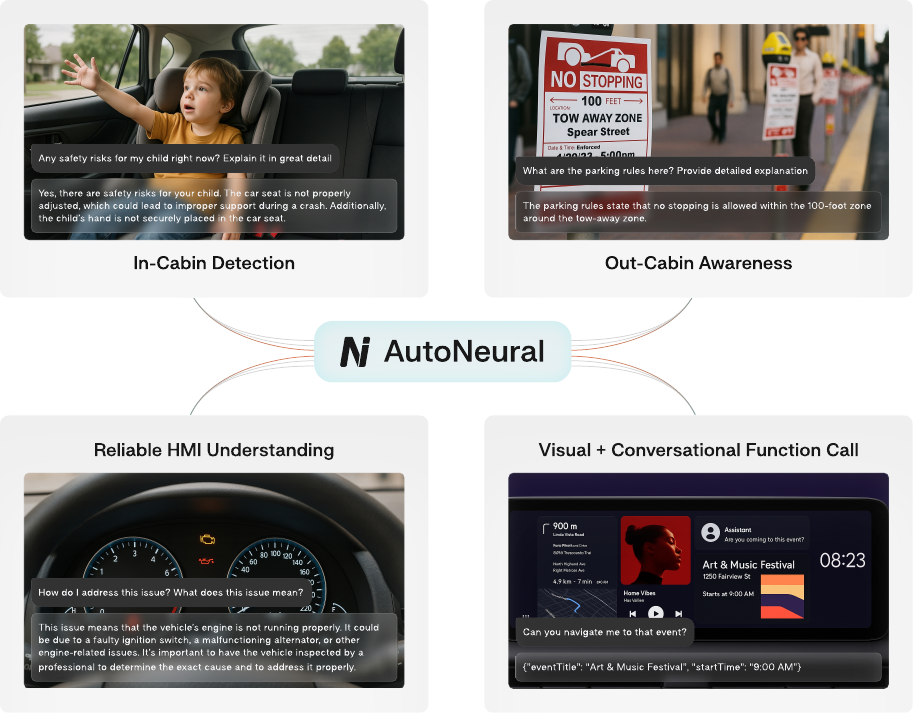}
\caption{Performance of AutoNeural after mixed-precision quantization (vision encoder: W8A16, language model: W4A16) and deployment on Qualcomm SA8295P NPU. Results reflect actual on-device execution, not PyTorch simulation, demonstrating stable accuracy and substantial latency improvements.}
\label{fig:model_functions}
\end{figure}

To validate the effectiveness of our architectural choices, we deployed the quantized AutoNeural model on the Qualcomm SA8295P NPU and measured real-world performance characteristics. Figure~\ref{fig:model_functions} presents the performance profile after mixed-precision quantization and on-device deployment, where the vision encoder operates at W8A16 (8-bit weights, 16-bit activations) and the language model backbone at W4A16 (4-bit weights, 16-bit activations). Critically, these measurements reflect actual NPU execution, not PyTorch simulation. The results demonstrate that our MobileNet encoder maintains stable accuracy under W8A16 quantization with minimal degradation, while the hybrid Transformer-SSM backbone achieves up to 60\% reduction in memory bandwidth during generation even at aggressive W4A16 precision. These on-device measurements confirm that our NPU-native co-design delivers substantial latency improvements—up to $14\times$ faster than ViT-Transformer baselines under comparable quantization settings—while preserving task accuracy.

\subsection{Cockpit AI and automotive applications}
Modern automotive cockpits are evolving from traditional human--machine interfaces into intelligent multimodal assistants that understand both visual and textual contexts~\cite{zhang2021pilot, kovalyov2024intelligent}. Edge-deployed multimodal models enable diverse interaction scenarios critical for driving safety and user experience. These include real-time interpretation of dashboard indicators and warning lights, with actionable recommendations based on vehicle sensor data. Scene understanding supports navigation assistance under challenging conditions, such as nighttime or adverse weather when visibility is compromised, while interpretation of traffic signs and road conditions through external cameras helps prevent navigation errors. The system also supports analysis of documents captured via in-cabin cameras, multimodal interaction combining gesture recognition with voice commands for hands-free operation, and driver state monitoring to detect fatigue or anxiety and provide timely interventions through environmental adjustments.

Unlike general-purpose VLMs, automotive cockpit AI faces stringent deployment constraints. Latency requirements demand responses within milliseconds to maintain driver attention and safety, while strict power budgets are dictated by vehicle electrical systems. Furthermore, cockpit deployments must maintain robustness across extreme environmental conditions including varying illumination from direct sunlight to nighttime darkness, temperature fluctuations ranging from sub-zero to high heat, mechanical vibration, and diverse user populations with different interaction patterns. These constraints necessitate NPU-native architectures that deliver consistent low-latency performance under quantized inference while maintaining high accuracy across automotive-specific visual and linguistic domains.

\section{Experiments}
\subsection{Training setup}

We utilize the Infinity-MM dataset~\cite{gu2024infinitymmscalingmultimodalperformance}, a large-scale, high-quality multimodal instruction dataset designed for comprehensive vision–language training. Infinity-MM contains approximately 44.8 million instruction-following samples spanning diverse visual understanding capabilities. The dataset is carefully curated to balance multiple task categories including general VQA covering everyday visual reasoning, object recognition, and scene understanding; document understanding with forms, receipts, scientific papers, and business documents; chart and diagram reasoning; OCR-centric tasks; multi-turn conversations; and specialized domains such as medical imaging, autonomous driving scenarios, and industrial inspection. Each sample consists of an image, an instruction sequence, and reference responses. The dataset emphasizes high-quality, detailed responses that require multi-step reasoning rather than simple label recognition, making it particularly suitable for training models that need to provide thorough explanations—critical for automotive safety applications where users must understand the model's reasoning process.

\subsubsection{Training Protocol}

Our training follows the four-stage curriculum prescribed by the Infinity-MM training protocol, progressing from basic image-text associations to complex instruction-following with increasing task diversity and image resolution.

\paragraph{Stage 1: Image-text association learning.}
In the first stage, we freeze both the vision encoder and the language model, training only the vision-language projection layer to establish basic image-text associations. We use high-quality image-caption pairs from Infinity-MM to optimize a next-token prediction objective where the model learns to generate captions conditioned on visual tokens. This stage runs for 1 epoch with a learning rate of 1e-3, using the AdamW optimizer with $\beta_1=0.9$, $\beta_2=0.95$, and a cosine learning rate schedule with warmup. The effective batch size is set to 512 with gradient accumulation.

\paragraph{Stage 2: General visual task training.}
In the second stage, we unfreeze all parameters, including both the vision encoder and the language model, and train on general visual understanding tasks such as object recognition, scene understanding, and basic visual question answering. We train with a learning rate of 1e-5 for 1 epoch, using the same optimizer configuration. For the first 600 training steps, a batch size of 16 is adopted to mitigate gradient explosion; the effective batch size is then set to 512 with gradient accumulation for the subsequent steps.

\paragraph{Stage 3: Instruction-specific fine-tuning.}
The third stage focuses on specialized instruction-following capabilities across the diverse task categories in Infinity-MM. We train on the full dataset comprising document understanding, chart reasoning, OCR-centric tasks, and multi-turn conversations. Following the Infinity-MM recipe, we employ task-specific mixture weighting that allocates 35\% to general VQA, 25\% to document understanding, 20\% to chart reasoning, 15\% to OCR tasks, and 5\% to multi-turn and specialized domains. The learning rate is 1e-5, and we train for 1 epoch with an effective batch size of 512. The training objective is standard autoregressive language modeling loss with next-token prediction.

\paragraph{Stage 4: Domain-specific quantization-aware fine-tuning and synthetic data integration.}
The final stage integrates both high-quality synthetically generated data and our custom automotive cockpit dataset to enhance the model's robustness and domain-specific capabilities. Following the Infinity-MM protocol, we incorporate synthetic samples that augment underrepresented task categories and challenging edge cases relevant to automotive deployment scenarios. Additionally, we introduce 0.2M automotive-specific samples that we collected and annotated, covering four critical cockpit AI tasks: AI Sentinel (56K samples for vehicle security monitoring), AI Greeter (50K samples for identity recognition and access control), AI Car Finder (44K samples for parking lot localization), and Safety Monitoring (50K samples for passenger egress and ingress safety). This automotive dataset was collected by approximately 400 volunteers across diverse demographics to ensure generalization across varied lighting conditions, vehicle types, and user populations. We use the quantization-aware training (QAT) method to fine-tune all parameters with a learning rate of 1e-5 for 1 epoch with an effective batch size of 512 using a mixture that balances 60\% synthetic data and 40\% automotive data. This stage employs careful quality filtering to ensure all samples maintain high annotation standards suitable for safety-critical automotive deployment.

\subsection{Comparison with other models}

We evaluate our NPU-native architecture against state-of-the-art vision–language models across five diverse multimodal benchmarks. Our comparison includes recent VLMs spanning different architectural families and parameter scales: InternVL2 (1B and 2B variants), representing ViT-Transformer architectures optimized for GPU deployment, and Qwen2-VL-2B and Qwen2.5-VL-3B-Instruct~\cite{wang2024qwen2}, demonstrating strong general-purpose multimodal capabilities. We report the performance of our AutoNeural model trained to Stage 3 of the Infinity-MM curriculum, which achieves the best balance between accuracy and training efficiency.

\paragraph{Benchmark suite.}
We evaluate on five benchmarks targeting distinct multimodal capabilities. \textbf{MMStar} measures core vision–language reasoning through multiple-choice visual questions requiring multi-step inference and commonsense knowledge. \textbf{HallusionBench} evaluates language hallucination and visual illusion in large vision--language models. \textbf{MathVista\_MINI} assesses mathematical reasoning over diagrams, charts, and geometric figures, testing the model's ability to extract quantitative information and perform numerical computations. \textbf{AI2D\_TEST} evaluates diagram understanding and scientific reasoning on annotated illustrations from textbooks and educational materials. \textbf{OCRBench} tests scene text recognition and document understanding across diverse text-in-image scenarios including natural scenes, documents, and signage. We report accuracy on each benchmark and compute the average score across all tasks to measure overall multimodal competence.

\paragraph{Results.}
Table~\ref{tab:comparison} presents the comprehensive benchmark comparison including ablation studies. AutoNeural achieves an average score of 60.75 across all benchmarks, demonstrating competitive performance with its NPU-optimized architecture and 1.47B parameter count. Notably, AutoNeural outperforms InternVL2-1B (55.96 average) and approaches InternVL2-2B (61.97 average) despite having 33\% fewer parameters. To isolate the impact of the vision encoder and the language model backbone, we conduct ablation experiments: comparing InternViT-Qwen (InternViT + Qwen2.5-1.5B, 63.08 average) with MobileNet-Qwen (MobileNetV5 + Qwen2.5-1.5B, 65.15 average) demonstrates that the MobileNet encoder is \textbf{not worse} than Vision Transformers while delivering 14$\times$ lower latency. Comparing MobileNet-Qwen (65.15) against AutoNeural (60.75) isolates the impact of the language backbone, showing that the Liquid AI LFM2-1.2B achieves a favorable trade-off: a modest accuracy reduction in exchange for 2.9$\times$ higher decode throughput and a 4$\times$ larger context length on NPU hardware.

\begin{table}[t]
\centering
\caption{Comparison of AutoNeural with state-of-the-art vision–language models including ablation studies across five multimodal benchmarks. All models are evaluated in full precision. Ablation rows (InternViT-Qwen, InternViT-Liquid, MobileNet-Qwen) systematically analyze vision encoder (InternViT vs. MobileNet) and language backbone (Qwen2.5-1.5B vs. LFM2-1.2B) contributions. All ablation models are trained by us using the same training strategy. All scores are accuracy percentages.}
\label{tab:comparison}
\footnotesize
\setlength{\tabcolsep}{2.5pt}
\begin{tabular}{l|c|c|c|c|c|c|c}
\toprule
\textbf{Model} & \textbf{Params} & \textbf{AI2D} & \textbf{Hallusion} & \textbf{MathVista} & \textbf{MMStar} & \textbf{OCR} & \textbf{Avg} \\
 & &\textbf{TEST} & \textbf{Bench} & \textbf{MINI} &  & \textbf{Bench} & \\
\midrule
InternVL2-1B & 0.94B & 64.09 & 54.26 & 40.30 & 45.47 & 75.70 & 55.96 \\
InternVL2-2B & 2.2B & 73.90 & 58.36 & 49.10 & 50.07 & 78.40 & 61.97 \\
Qwen2-VL-2B & 2.2B & 74.64 & 62.04 & 47.10 & 47.80 & 80.90 & 62.50 \\
Qwen2.5-VL-3B-Instruct & 3.75B & 81.51 & 63.62 & 62.00 & 56.73 & 82.70 & 69.31 \\
\midrule
InternViT-Qwen & 1.86B & 77.17 & 55.31 & 57.50 & 54.93 & 72.50 & 63.08 \\
InternViT-Liquid & 1.49B & 72.73 & 54.26 & 52.60 & 50.87 & 66.00 & 59.29 \\
MobileNet-Qwen & 1.84B & 78.63 & 57.94 & 60.60 & 55.07 & 73.50 & 65.15 \\
\midrule
\textbf{AutoNeural} & \textbf{1.47B} & \textbf{73.80} & \textbf{56.05} & \textbf{53.10} & \textbf{49.40} & \textbf{71.40} & \textbf{60.75}\\
\bottomrule
\end{tabular}
\end{table}

\subsection{Latency and quantization analysis}

\begin{figure}[t]
\centering
\includegraphics[width=0.75\textwidth]{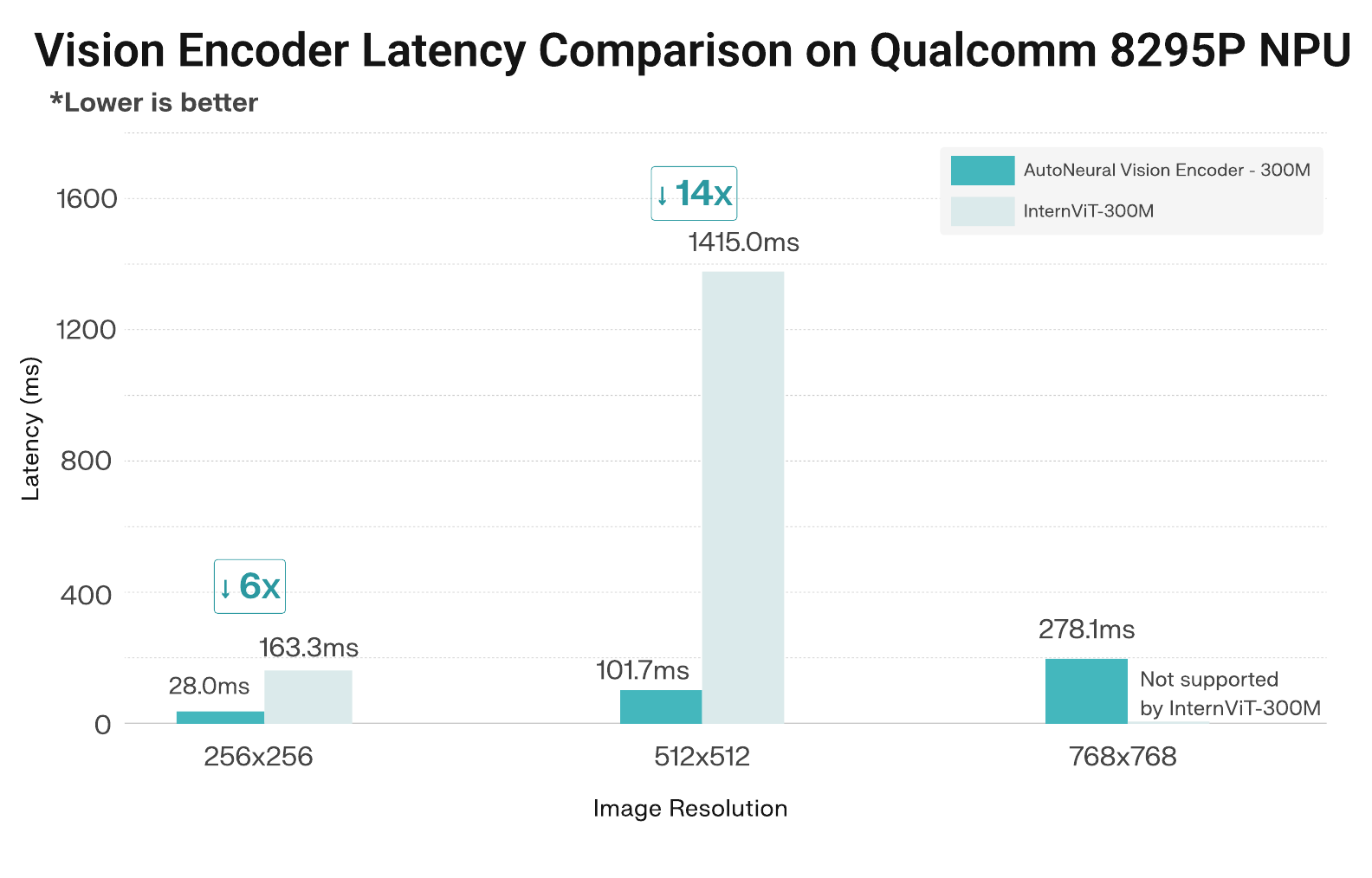}
\caption{Vision encoder latency comparison on Qualcomm SA8295P NPU across three input resolutions. AutoNeural's MobileNet-based encoder achieves 5.8$\times$ speedup at 256$\times$256, 14$\times$ speedup at 512$\times$512, and successfully processes 768$\times$768 images in real time, while InternViT-300M exceeds the NPU memory capacity. Lower latency is better.}
\label{fig:latency_comparison}
\end{figure}

While accuracy metrics demonstrate competitive performance, the primary motivation for our NPU-native architecture is to achieve substantially lower latency under edge deployment constraints. We conducted on-device latency measurements on the Qualcomm SA8295P NPU to quantify the real-world speedup of our MobileNet-based vision encoder compared to ViT-based alternatives. Figure~\ref{fig:latency_comparison} presents vision encoder latency across three input resolutions: 256$\times$256, 512$\times$512, and 768$\times$768 pixels. Both models are deployed with W8A16 quantization (8-bit weights, 16-bit activations) to ensure a fair comparison under identical precision constraints. Our AutoNeural vision encoder (300M parameters) is compared against InternViT-300M, a Vision Transformer encoder with a comparable parameter count.

The results reveal dramatic latency advantages across all tested resolutions. At 256$\times$256 resolution, AutoNeural achieves 28.0ms inference time compared to 163.3ms for InternViT-300M, yielding a 5.8$\times$ speedup. At 512$\times$512, the gap widens substantially: AutoNeural processes images in 101.7ms while InternViT-300M requires 1415.0ms—a 14$\times$ speedup. Most critically, at the native 768$\times$768 resolution used in our full VLM system, AutoNeural maintains real-time performance at 278.1ms per image, whereas InternViT-300M fails to execute due to memory constraints on the NPU hardware. This inability to support high-resolution inputs fundamentally limits ViT-based architectures for automotive cockpit applications, where detailed visual understanding of dashboard indicators, traffic signs, and scene context requires processing at sufficient resolution.

The latency gap stems from fundamental architectural differences. Vision Transformers employ global self-attention with quadratic complexity in the number of patches, generating large intermediate activation tensors. In contrast, our MobileNet encoder's depthwise separable convolutions maintain bounded activation footprints, enable efficient NPU operator fusion, and exhibit predictable memory access patterns that maximize on-chip buffer utilization. The MSFA module further reduces latency by processing only two downsampled feature maps rather than maintaining full-resolution activations throughout the network. These results validate that NPU-native architectural choices are essential for real-time multimodal intelligence at the edge, where latency constraints are as critical as accuracy metrics.

\paragraph{End-to-end system performance.}
To evaluate the complete system under realistic deployment conditions, we compare AutoNeural-VL against InternVL 2B as a baseline solution on the Qualcomm SA8295P NPU. Table~\ref{tab:system_benchmark} presents comprehensive metrics including TTFT, maximum image resolution, signal-to-quantization-noise ratio (SQNR), quantization error (RMS Error), LLM Perplexity, decode throughput, and context length. AutoNeural-VL achieves full vision support with 768$\times$768 resolution compared to InternVL 2B's 448$\times$448 limitation, enabling detailed visual understanding crucial for automotive applications. Critically, AutoNeural-VL delivers 14$\times$ faster TTFT ($\sim$100ms vs. $\sim$1.40s for a 512$\times$512 image), demonstrating the dramatic latency improvements enabled by our NPU-native architecture. The MobileNet encoder maintains superior quantization robustness with 7$\times$ lower RMS error (0.562\% vs. 3.98\%) and 17 dB higher SQNR (45 dB vs. 28 dB), validating that it is fundamentally more stable under aggressive quantization. To further validate the language model's quantization robustness, we measure perplexity on a held-out test set: the Liquid AI 1.2B backbone exhibits minimal degradation from 21.13 (FP16) to 21.47 (W4A16), corresponding to only 1.6\% perplexity increase under aggressive 4-bit weight quantization. This stability confirms that the hybrid Transformer-SSM architecture maintains text generation quality while achieving substantial memory and compute savings. Additionally, AutoNeural-VL achieves 2.9$\times$ higher decode throughput ($\sim$44 tok/s vs. $\sim$15 tok/s) and supports a 4$\times$ larger context length (4096 vs. 1024), enabling more complex multimodal reasoning. These results demonstrate that NPU-native co-design delivers substantial improvements across all critical deployment metrics—latency, quantization robustness, throughput, and context capacity—making real-time automotive cockpit AI practical.

\begin{table}[t]
\centering
\caption{End-to-end system performance comparison between AutoNeural-VL (vision encoder: W8A16, language model: W4A16) and the InternVL 2B baseline (vision encoder: W8A16, language model: W4A16) on the Qualcomm SA8295P NPU. AutoNeural-VL achieves 14$\times$ faster TTFT, 3$\times$ faster decoding, and 7$\times$ lower quantization error, and supports a 4$\times$ larger context length while maintaining higher throughput and resolution.}
\label{tab:system_benchmark}
\footnotesize
\setlength{\tabcolsep}{6pt}
\begin{tabular}{l|c|c}
\toprule
\textbf{Metric} & \textbf{InternVL 2B} & \textbf{AutoNeural-VL} \\
\midrule
Vision encoder latency (1 image, 512$\times$512) & $\sim$1.4 s & $\sim$100 ms \\
Max Image Size & 448$\times$448 & 768$\times$768 \\
SQNR & 28 dB & 45 dB \\
RMS Error & 3.98\% & 0.562\% \\
LLM Perplexity (FP16 $\rightarrow$ W4A16) & - & 21.13 $\rightarrow$ 21.47 \\
Decode Throughput & $\sim$15 tok/s & $\sim$44 tok/s \\
Context Length & 1024 & 4096 \\
\bottomrule
\end{tabular}
\end{table}

\paragraph{Future work.}
The benchmark results presented in this work are conducted under full precision conditions. While we provide quantization robustness metrics through SQNR measurements for the vision encoder and perplexity analysis for the language model, comprehensive in-vehicle evaluation under real deployment conditions and the exploration of better quantization approaches remain essential future work. This includes validating quantized model performance across diverse driving scenarios, extending benchmarks to additional NPU platforms, conducting end-to-end quality assessment in production automotive environments, and exploring automated architecture search for hardware-specific optimization.

\section{Conclusion}

We presented an NPU-native vision–language model architecture that fundamentally rethinks both vision encoding and language modeling for efficient edge deployment. By replacing the standard ViT-Transformer paradigm with a MobileNet-based encoder and a hybrid Transformer-SSM backbone, we address the dual bottlenecks of quantization brittleness and memory I/O that plague existing VLM architectures on NPUs. Empirical evaluation across standard multimodal benchmarks reveals that our architecture delivers up to $7\times$ lower quantization error and $14\times$ lower end-to-end latency compared to ViT-Transformer baselines under the same NPU precision constraints. The MobileNet encoder achieves stable INT8/16 inference with minimal quantization degradation, while the hybrid Transformer-SSM backbone reduces memory I/O by up to 60\% during autoregressive generation. Real-world deployment on the Qualcomm SA8295P NPU validates real-time responsiveness in automotive scenarios.

Our work demonstrates that NPU-native co-design is essential for robust, low-latency multimodal intelligence at the edge. The conventional approach of scaling resolution and model capacity exhibits diminishing returns on NPUs, where quantization brittleness and memory bandwidth become the primary bottlenecks. By carefully selecting operators that align with NPU execution models—depthwise separable convolutions for vision, SSMs for language—we unlock the full potential of edge accelerators. This principle extends beyond vision–language models to any multimodal architecture targeting resource-constrained deployment. Future work includes exploring automated architecture search conditioned on hardware profiles, mixed-precision quantization schemes, and validation across diverse edge accelerators to strengthen generalizability.

\medskip
{\small
\bibliographystyle{unsrtnat}
\bibliography{citation}
}


\end{CJK*}
\end{document}